\documentclass{IEEEoj}
\usepackage{cite}
\usepackage{amsmath,amssymb,amsfonts}
\usepackage{algorithmic}
\usepackage{graphicx,color}
\usepackage{textcomp}
\usepackage{xr}
\usepackage{cleveref}
\usepackage{booktabs}
\usepackage{soul}
\renewcommand{\ul}[1]{#1}

\def\BibTeX{{\rm B\kern-.05em{\sc i\kern-.025em b}\kern-.08em
    T\kern-.1667em\lower.7ex\hbox{E}\kern-.125emX}}
\makeatletter
\def\ps@headings{\def\@oddhead{\vbox{\vspace{17pt}\hsize\textwidth\hbox{\rfxfont\rightmark\hfill}\hfill\par
\smallskip\noindent\hbox to \textwidth{\vrule width\textwidth height.3pt depth0pt}}}%
\def\@evenhead{\vbox{\vspace{17pt}\hsize\textwidth\hfill\hbox{\hfill\rhfont\leftmark}\par
\smallskip\noindent\hbox to \textwidth{\vrule width\textwidth height.3pt depth0pt}}}%
\def\@oddfoot{\hfill\rffont\thepage}\def\@evenfoot{\rffont\thepage\hfill}}
\def\ps@plain{\def\@oddhead{\vbox{\vspace{17pt}\hsize\textwidth\hbox{\rhfont\leftmark\hfill}\hfill\par
\smallskip\noindent\hbox to \textwidth{\vrule width\textwidth height.3pt depth0pt}}}%
\def\@evenhead{\vbox{\vspace{17pt}\hsize\textwidth\hfill\hbox{\hfill\rhfont\leftmark}\par
\smallskip\noindent\hbox to \textwidth{\vrule width\textwidth height.3pt depth0pt}}}%
\def\@oddfoot{\hfill\rffont\thepage}\def\@evenfoot{\rffont\thepage\hfill}}
\makeatother
\AtBeginDocument{\definecolor{ojcolor}{cmyk}{1,0.45,0,0.18}%
\definecolor{ojcolor2}{cmyk}{0,0.91,0.81,0.19}%
\pagestyle{headings}\thispagestyle{plain}}
\begin{document}
\receiveddate{XX Month, XXXX}
\reviseddate{XX Month, XXXX}
\accepteddate{XX Month, XXXX}
\publisheddate{XX Month, XXXX}
\currentdate{XX Month, XXXX}
\doiinfo{JXCDC.2023.3238030}

\title{\textcolor{ojcolor2}{X-TIME: accelerating large tree ensembles inference for tabular data with analog CAMs}}

\author{GIACOMO PEDRETTI, MEMBER, IEEE, 
        JOHN MOON, 
        PEDRO BRUEL, 
        SERGEY SEREBRYAKOV,
        RON M. ROTH, FELLOW, IEEE,
        LUCA BUONANNO,
        ARCHIT GAJJAR,
        LEI ZHAO,
        TOBIAS ZIEGLER,
        CONG XU, 
        MARTIN FOLTIN, 
        PAOLO FARABOSCHI, FELLOW, IEEE,
        JIM IGNOWSKI, SENIOR MEMBER, IEEE,
        CATHERINE E. GRAVES}
\affil{Artificial Intelligence Research Lab (AIRL), Hewlett Packard Labs, Milpitas, CA 95035 USA}
\affil{Artificial Intelligence Research Lab (AIRL), Hewlett Packard Labs, Fort Collins, CO 80528 USA}
\corresp{CORRESPONDING AUTHOR: Giacomo Pedretti (e-mail: giacomo.pedretti@hpe.com).}
\authornote{This work was funded by Army Research Office under Agreement W911NF2110355.}


\begin{abstract}
Structured, or tabular, data is the most common format in data science. While deep learning models have proven formidable in learning from unstructured data
such as images or speech, they are less accurate than simpler approaches when learning from tabular data. 
In contrast, modern tree-based Machine Learning (ML) models shine in extracting relevant information from structured data. An essential requirement in data science is to reduce
model inference latency in cases where, for example, models are used in a closed loop with simulation to accelerate scientific discovery. However, the hardware acceleration
community has mostly focused on deep neural networks and largely ignored other forms of machine learning. Previous work has described the use of an analog content addressable memory (CAM) component for efficiently mapping random forests. In this work, we develop an analog-digital architecture that implements a novel increased precision analog CAM and a programmable chip for inference of state-of-the-art tree-based ML models, such as XGBoost, CatBoost, and others. Thanks to hardware-aware training, X-TIME reaches state-of-the-art accuracy and 119$\times$ higher throughput at 9740$\times$ lower latency with $>150\times$ improved energy efficiency compared with a state-of-the-art GPU \ul{for models with up to 4096 trees and depth of 8}, with a 19W peak power consumption.
\end{abstract}


\maketitle

\section{INTRODUCTION}

\IEEEPARstart{E}{xtracting} relevant information from structured (i.e., tabular) data is the foundation of practical data science. 
Tabular data consists of a matrix of samples organized in rows, each of which has the same set of features in its columns, and is commonly used for medical, financial, scientific, and networking applications, enabling efficient search operations on large-scale datasets\ul{, making them one of the most common sources of data in the data science industry. Moreover, data collected in real-time for heterogenous sources, for example during scientific experiments, are organized in tabular data and with fast processing may lead to in-the-loop intelligent control.}

While deep learning has shown impressive improvements in accuracy for unstructured data, such as image recognition or machine translation, tree-based machine learning (ML) models still outperform neural networks in classification and regression tasks on tabular data~\cite{grinsztajn_why_2022}, \ul{including modern Transformers for tabular data}~\cite{huang2020tabtransformer}.
This difference in model accuracy is due to robustness to uninformative features of tree-based models, bias towards a smooth solution of neural networks models, and in general the need for rotation variant models where the data is rotation variant itself, as in the case of tabular data~\cite{grinsztajn_why_2022}.
With high model accuracy, ease of use, and explainability~\cite{lundberg_local_2020} tree-based ML models are greatly preferred by the data science community~\cite{kaggle_state_2021}, with $>74\%$ of data scientists preferring to use these models while $<40\%$ choose neural networks in a recent survey. 

However, tree-based models have garnered little attention in the architecture and custom accelerator research fields, particularly compared to deep learning. 
While small models are easily executed, performance issues for larger tree-based ML models arise due to limited inference latency both on CPU and GPUs~\cite{xie_tahoe_2021,he_booster_2022} originating from irregular memory accesses and thread synchronization issues. 
As large tree-based models ($>$1M nodes) are finding performance limits in existing hardware, they are demonstrating increasingly compelling uses in practical environments where high throughput and moderate latency are desired. 
For example, the winner of a recent IEEE-CIS Fraud detection competition used a tree-based ML model with~20M nodes~\cite{ieee_cis}.
Moreover, reducing inference latency is critical in scientific applications where for use in real-time processing, model latency must be $\sim$100 ns~\cite{summers_fast_2020}. 
Similar low latency requirements for model inference arise in a real-time in-the-loop decision, data filtering systems, or streaming data scenarios, which are common across financial, medical, and cybersecurity domains~\cite{gajjar_faxid_2022,ieee_cis} where tabular data and tree-based ML also dominate. 
Thus, these models are reaching sizes and complexity levels for meaningful real-world tasks, as they are hitting limits on inference latency and throughput in current systems. 
Recently, custom accelerators for tree-based ML inference have been developed based either on FPGA \cite{summers_fast_2020,gajjar_faxid_2022} or custom ASICs \cite{tzimpragos_racetree_2019,he_booster_2022}. 
A common architecture in most of the accelerators mentioned above implements look-up tables (LUTs) storing the attributes of a decision tree (DT) accessed multiple times during inference, with reads based on decision paths taken during the tree traversal. 
Along with low latency inference, an efficient hardware representation of trees is necessary to implement the required very large models.

One promising new architectural approach for improving tree-based model inference is in-memory computing, where data are stored and processed in the same location. 
This approach eliminates access, transfer, and processing of data in a separate compute unit~\cite{zidan_future_2018}, reducing latency.
Previous research has shown that multiplication functions can efficiently be performed in crossbar architectures~\cite{shafiee_isaac_2016}.
However, crossbar architectures are limited to matrix operations, which are not useful
for inference on tree-based models. 
Another class of in-memory computing primitives is Content Addressable Memories (CAMs)~\cite{pagiamtzis_content-addressable_2006,graves_-memory_2022}, high-speed and highly parallel circuits that search for a given input in a table of stored words and return the matched search input's location. 
CAMs use relatively cheap peripherals, e.g. sense amplifier circuits rather than expensive ADCs required by crossbar arrays for in-memory computing operations.
More recently, analog CAMs have been proposed~\cite{li_analog_cam_2020} which leverage the analog states in resistive switching memories (RRAM) \cite{ielmini_resistive_2016}. 
An analog CAM performs the comparisons of stored (analog) ranges with (analog) inputs, by computing if the input is within the range. 
Given that DTs are a set of conditional branches, analog CAMs naturally map DTs and perform highly parallel inference of decision forests~\cite{pedretti_tree-based_2021,yin2024deep}. 
Our previous work showed the potential of this approach but was limited to component-level estimation, 4-bit precision, and random forest models~\cite{pedretti_tree-based_2021}, \ul{while the full architecture study in this work - which includes designs for higher precision Analog CAM operation, study of peripheral circuits and builds in support for multiple ML models convincingly demonstrates the feasibility of the initial concept.} 

This work presents X-TIME, an in-memory engine for accelerating tree-based ML models with analog CAMs. 
In this work we demonstrate the need for a vertical architecture design, 8-bit operation, and support for multiple tree-based ML models, overcoming the limitations of previous work~\cite{pedretti_tree-based_2021}. 
The major contributions of this paper compared to previous work~\cite{pedretti_tree-based_2021} are \ul{at the circuit level}(1) a novel analog CAM cell design and operation to perform decision tree inference at doubled (i.e. 8-bit) resistive memory precision (i.e. 4-bit), (2) \ul{a complete design of the core and network operation, and at the architecture level} (3) an end-to-end design targeting a large scale accelerator, (4) a compiler (X-TIME-C) supporting the mapping of multiple tree-based ML models to the HW starting from established intermediate representation \cite{treelite}, \ul{and at the co-design level } (5) an HW-aware model training algorithm for reaching state of the art accuracy and (6) a complete simulation framework to benchmark and rigorous characterization comparison against \ul{NVIDIA V100} GPUs, FPGAs, and conventional digital accelerators.
Our simulated evaluation shows up to 119$\times$ increased throughput at 9740$\times$ decreased latency with $>150\times$ improved energy efficiency in a single 19W chip. 

\section{BACKGROUND}
\begin{figure}
\centering
\includegraphics[width=0.8\linewidth]{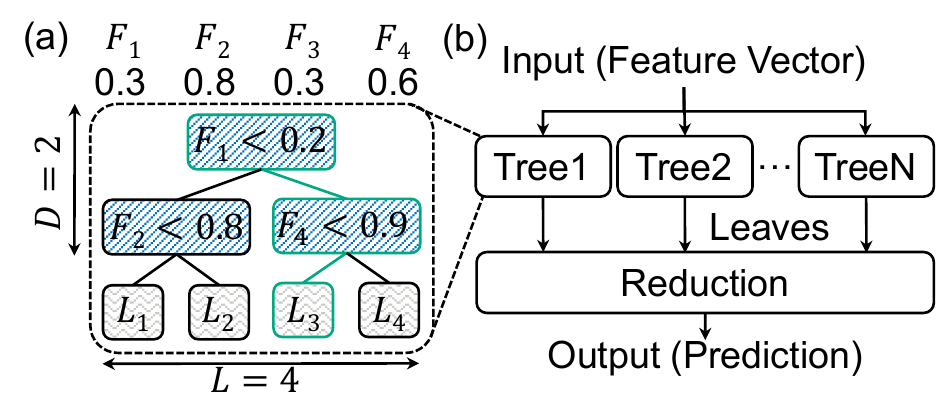}
\caption{Schematic illustration of (a) a decision tree and (b) a decision tree ensemble.} 
\label{fig1}
\end{figure}
\subsection{Tree-based Machine Learning}
A DT is a regression and classification ML model consisting of a set of \textit{nodes} arranged in a tree-like structure. 
Each node can have multiple children nodes, but the most common case is a \textit{binary tree}, where each node has only two children. The first node is usually called the \textit{root}. 
Each node $j$ consists of a conditional branch where a single input feature element $f_i$ is compared with a trainable threshold $T_j$.
The exception to this is the last node, or \textit{leaf}, which stores the model's prediction for a given input, and which can be a class (classification) or a scalar value (regression).
A balanced tree can be defined by its \textit{depth} $D$, which is equal to the maximum number of nodes in a root-to-leaf traversal. 
Fig.~\ref{fig1}(a) shows an example of a DT with $L=4$ leaves and $D=2$.

Multiple DTs can be trained together as an ensemble for increased accuracy in classification and regression tasks. 
Fig.~\ref{fig1}(b) shows a conceptual representation of a DT ensemble, where multiple trees are executed in parallel and their output combined with a \textit{reduction} operation yields the model's final result. 
Examples of DT ensembles model include Random Forests (RF)~\cite{biau_random_2016}, eXtreme Gradient Boosting (XGBoost)~\cite{chen_xgboost_2016},  and Categorical Boosting 
(CatBoost)~\cite{prokhorenkova_catboost_2018}. 
In the case of RF, the reduction operation consists of a majority voting, while in all the others leaf values of each DT are summed.

While training DT models is relatively fast when compared to, for example, a large deep neural network, the computations involved in inference are very irregular and not particularly suitable for CPUs and GPUs~\cite{he_booster_2022}, becoming inefficient for very large models, with~$>1$~M nodes \cite{ieee_cis}. 
While CPUs can run irregular control paths fairly well, their parallelism is limited. 
GPUs have a very large parallelism, but suffer from the irregularity of memory access patterns, as explained in detail in Supplementary Information \ref{supp-si:section:gpu_inference}.

\begin{figure}
\centering
\includegraphics[width=0.99\linewidth]{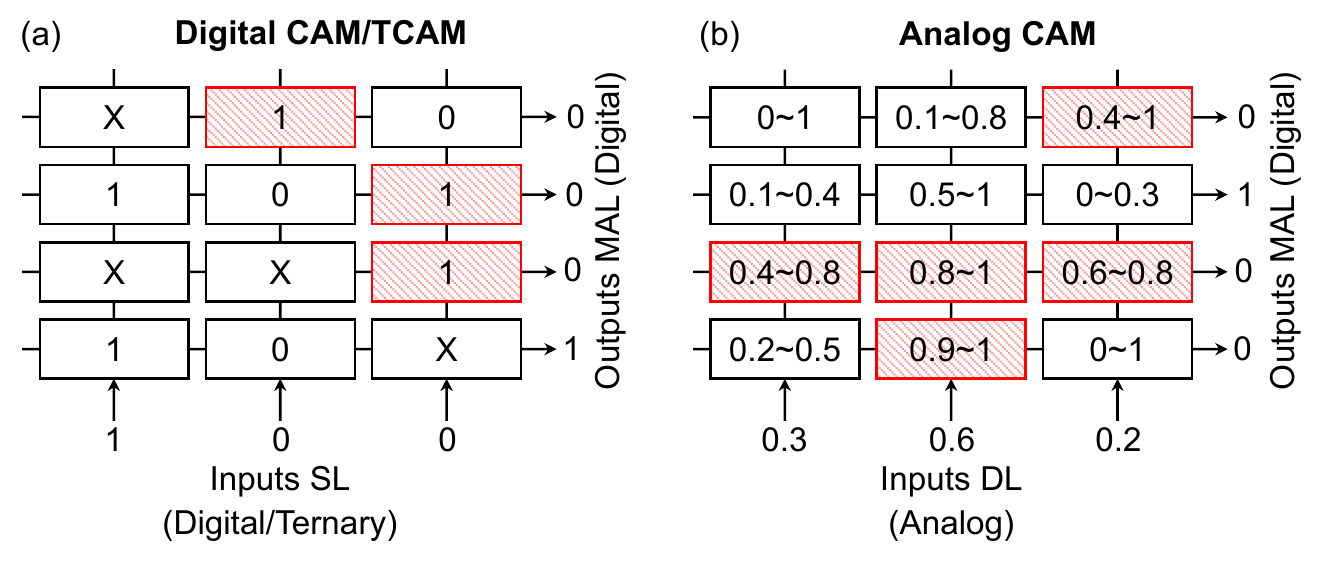}
\caption{Comparison of (a) a conventional CAM/TCAM and (b) an analog CAM} 
\label{fig2}
\end{figure}
\subsection{Analog Content Addressable Memory}
CAMs are specialized memory devices that search a table of stored words (rows) or keys in parallel for a given input query. 
CAMs output a one-hot vector where a high value corresponds to a match between the input data query and stored row word key~\cite{pagiamtzis_content-addressable_2006,graves_-memory_2022}. 
In conventional CAMs, this output vector is converted into a single address by a priority encoder. 
Each CAM cell functions as both memory and bit comparison, and the input query is compared to each CAM word simultaneously to produce a high throughput, low latency few-cycle compare operation at the cost of high power and area. 
In addition to binary (0 and 1) values, ternary CAMs (TCAMs) have a third wildcard value, \textit{X} (or \textit{don't care}), which can be used to compress the stored CAM table or perform partial match searches. 
When an \textit{X} is stored or searched, that cell matches irrespective of whether the searched or stored value is 1 or 0. 
Fig.~\ref{fig2}(a) shows a schematic illustration of a conventional CAM/TCAM circuit, where input queries are applied on the column Search Lines (SLs) and outputs are read along the row Match Lines (MALs). 
Essentially, an XNOR operation occurs between the query values and stored cell values in the corresponding column. 
The MAL $j_{th}$ row output is produced by a bit-wise AND operation between cells along the row and returns a match (high) or mismatch (low). 
The red boxes in Fig.~\ref{fig2}(a) correspond to mismatches between input and stored value.
In practice, MALs are initially pre-charged high and a mismatch in a cell activates a pull-down transistor connected to MAL to discharge it, such that a match is returned only if all pull-down transistors remain off during a search operation.

TCAMs are used ubiquitously in networking applications that require high-throughput and low-latency operations, for example for packet classification and forwarding, and access control lists~\cite{pagiamtzis_content-addressable_2006}. 
While offering high performance, conventional SRAM-TCAMs are power and area-hungry and require 16 transistors. 
Recently, there has been significant interest in developing TCAMs with RRAM devices ~\cite{guo_resistive_2011,graves_memory_2020,li_sapiens_2021} with compact size and low-power operation.
As programming these devices is relatively slow, emerging TCAMs have targeted in-memory computing applications~\cite{ielmini_-memory_2018, graves_-memory_2022} where
stored value updates are infrequent. 

Beyond direct digital memory operation, RRAM devices have also demonstrated ranges of stable analog states of $>16$ levels~\cite{sheng2019low}. 
Leveraging this analog state capability, recent work proposed and demonstrated an analog CAM circuit design and operation~\cite{li_analog_cam_2020} (conceptual representation in Fig~\ref{fig2}(b)). 
In an analog CAM, \textit{ranges} are stored in each cell, an analog input is applied vertically along columns and a match is returned along rows if the inputs are all within the stored range of each cell in a row. An analog CAM row implements
\begin{equation}
    \text{MAL}_i = \bigwedge_j{(T_{L,ij}\leq q_j<T_{H,ij})}\text{,}
\end{equation}
where $q$ is the query input vector applied to Data Line (DL), and $T_{L}$ and $T_{H}$ are the stored range thresholds for the CAM cell range lower and upper bounds, respectively. 
In the analog CAM circuit, no analog-to-digital conversion is required, eliminating an expensive component which is a limiting bottleneck of other analog in-memory computing primitives~\cite{rekhi_analogmixed-signal_2019}. 
Multiple analog CAM circuits have been realized using different technologies such as RRAM devices~\cite{li_analog_cam_2020} and ferroelectric memory devices~\cite{ni_ferroelectric_2019}. 
In such circuits, two analog memory elements represent the lower and upper bound thresholds. 
Analog CAMs have been shown to map CAM tables more efficiently~\cite{li_analog_cam_2020}, as well as implement novel functions, such as computing distances for memory-augmented neural networks~\cite{ni_ferroelectric_2019,li_sapiens_2021}, decision tree inference~\cite{pedretti_tree-based_2021} and pattern learning \cite{pedretti_differentiable_correct}.

\begin{figure}
\centering
\includegraphics[width=0.6\linewidth]{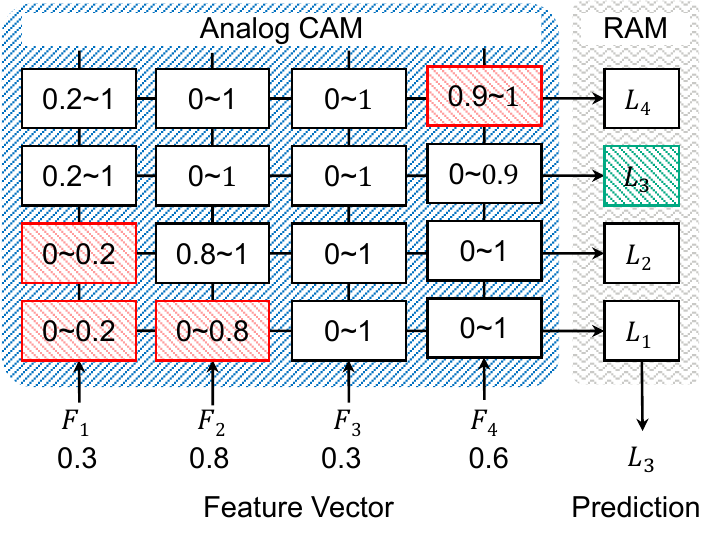 }
\caption{DT example from Fig.\ref{fig1}(a) with nodes mapped in an analog CAM and leaves mapped into a RAM.} 
\label{fig3}
\end{figure}
\subsection{Mapping tree models to analog CAMs}
Fig.~\ref{fig3} shows the mapping of the decision tree in Fig.~\ref{fig1}(a) to the Analog CAM.
Each row represents a branch of the tree and each column corresponds to a different feature of the input vector $q$.
Each node is mapped in the column corresponding to the feature used in the node's comparison.
The comparison's threshold value is programmed in the analog CAM's lower and/or upper bound, according to the direction, left or right, taken by the comparison's result for the root-to-leaf path. 
A \textit{don't care} symbol is used if a feature is missing, by programming the full range (0 to 1 in our example) in the analog CAM cell. 
The feature vector (query) is applied on the DL and the output on the MAL is high for the matched branch, i.e. the traversed root-to-leaf path, in a single CAM operation. 
This mapping requires the number of CAM rows to match the number of tree leaves, and the number of CAM columns to match the number of features $N_{feat}$. 
MALs can be connected directly to the Word Lines (WLs) of a conventional SRAM storing leaf values to return the model's prediction. 
Effectively, the analog CAM traverses root-to-leaf paths, and the RAM retrieves leaf values.

Previous results of this approach have shown significant speedups for RF inference~\cite{pedretti_tree-based_2021} and reduced energy consumption compared to both GPU and state-of-art ASICs~\cite{kang_194-njdecision_2018}, although with limited architecture investigation for scaling and no flexibility to accommodate different dataset or model types. 
In this work, we present a complete architecture development and study for handling multiple DT ensemble algorithms such as RF, XGBoost, and CatBoost, and learning tasks such as regression, binary classification, and multi-class classification with the same architecture. 
In addition, we extend the use of the previously presented analog CAM~\cite{li_analog_cam_2020,pedretti_tree-based_2021} circuit to support higher precision data by developing a flexible connectivity strategy and bit-arithmetic refactoring. 
We compare the performance of our architecture on state-of-the-art tabular dataset ML tasks against GPU, FPGA, and digital ASIC, showing a
large improvement in latency and throughput while using a single chip with limited power consumption.

\section{Hardware-aware model training}\label{sec:training}
To ensure a targeted and relevant accelerator design, we explore the HW algorithm co-design space by assessing model types, sizes, and accuracy requirements in conjunction with different analog CAM HW and architecture parameters. First, we consider the limitations of the previously presented RF accelerator based on analog CAM \cite{pedretti_tree-based_2021}.
The accelerator was limited to 4-bit precision, essentially due to the programming accuracy of ReRAMs, and it was benchmarked on a single dataset.
However, different tree-based ML models would perform differently for different tasks, such as binary or multiclass classification or regression, and different datasets.
The 4-bit operation might not be enough to capture more complex datasets that would require a larger number of \textit{bins} for each feature.
Finally, the maximum depth of each tree and the number of trees, corresponding to the number of words in the analog CAM and the number of analog CAM arrays, should be optimized on a variety of datasets to ensure a general well-performing accelerator.
Thus we performed an in-depth characterization of tree-based ML models, namely RF, XGBoost, and CatBoost, performance on several datasets to understand (1) the necessity for mapping diverse models to the accelerator, (2) the size of each analog CAM macro and total number of macros in the accelerator and (3) the required bit precision for the analog CAM operation.
\begin{table*}[h]
  \center
  \caption{Datasets and models characterization}\label{table:datasets}
  \begin{tabular}{lrrrrrrrrrrrr}
  \toprule
    \textbf{Dataset} & \textbf{ID} &\textbf{Task} & \textbf{Samples} & $\pmb{N_{feat}}$ & $\pmb{N_{classes}}$ & \textbf{Model} & $\pmb{N_{trees}}$ & $\pmb{N_{leaves,max}}$\\
\midrule
    Churn modelling\cite{churn_modelling} & 1 & Binary Classification & 10000 & 10 & 2 & CatBoost & 404 & 256\\

    Eye movements\cite{salojarvi2005inferring}& 2 & Multiclass Classification & 10936 & 26 & 3 & XGBoost & 2352 & 256\\

    Forest cover type\cite{blackard1999comparative}& 3 & Multiclass Classification & 581012 & 54 & 7 & XGBoost & 1351 & 231\\

    Gas concentration\cite{vergara_chemical_2012}& 4 & Multiclass Classification & 13910 & 129 & 6 & Random Forest & 1356 & 217\\
    
    Gesture phase segmentation\cite{wagner2014gesture}& 5 & Multiclass Classification & 9873 & 32 & 5 & XGBoost & 1895 & 256\\

    Telco customer churn\cite{telco_customer_churn}& 6 & Binary Classification & 7032 & 19 & 2 & XGBoost & 159 & 4\\

    Rossmann stores sales\cite{rossmann_store_sales}& 7 & Regression & 610253 & 29 & N/A & XGBoost & 2017 & 256\\
\bottomrule
  \end{tabular}
\end{table*}
We perform a survey on the size of typical tabular datasets \cite{grinsztajn_why_2022}, finding that usually $N_{feat}<100$ and often feature pre-processing reduces the number of $N_{feat}$ significantly before applying inputs to the decision tree ensemble. 
Nevertheless, we include an outlier dataset with $N_{feat} = 130$\cite{vergara_chemical_2012} to show how to handle larger input vector sizes with input vector segmentation, as will be described later. 
Table~\ref{table:datasets} shows a characterization of the selected datasets.
\begin{figure}[h!]
\centering
\includegraphics[width=0.99\linewidth]{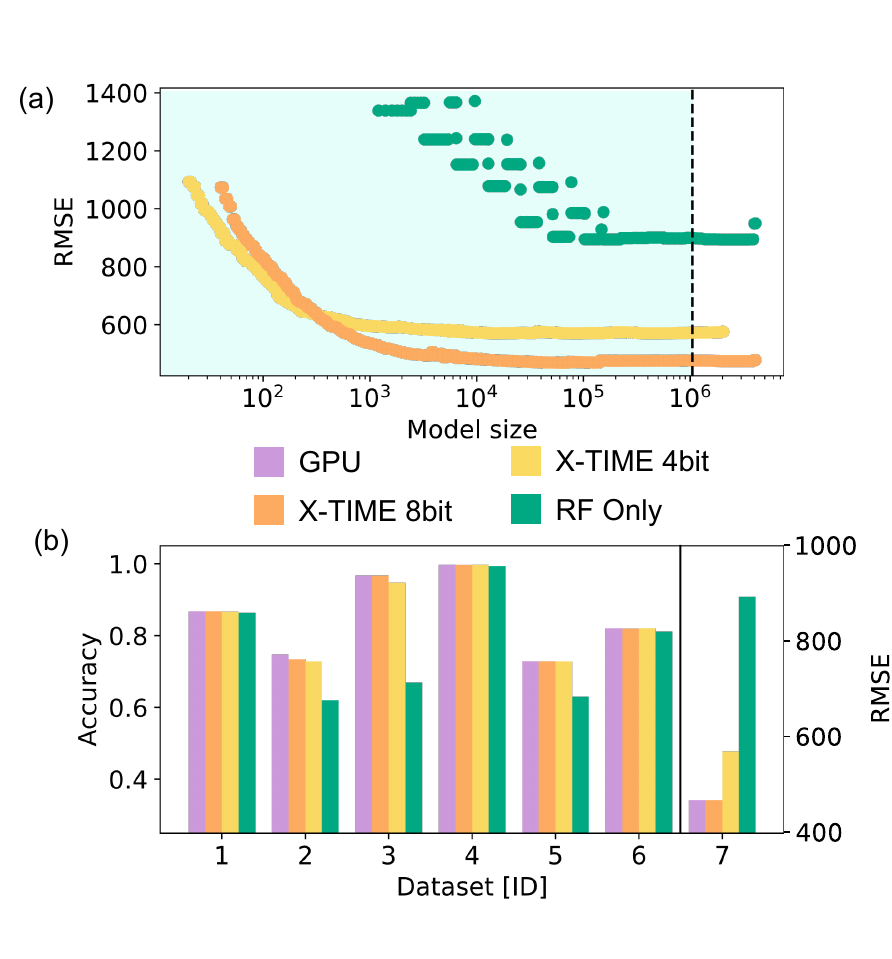}
\caption{(a) RMSE as a function of model size in the case of Rossmann stores sales\cite{rossmann_store_sales} for multiple configurations.(b) Best accuracy of each dataset for multiple configurations.} 
\label{fig:dse}
\end{figure}

First, we trained all the models optimizing the hyperparameters without limitations, similar to what would happen if training the model for the GPU.
Then, we studied the impact of bit precision. 
Given that from implementing 4-bit to 8-bit operation the analog CAM area would roughly double, as explained in Section \ref{section:precision}, we doubled the number of available trees in that case.
Finally, we limited the training to RF to demonstrate that it underperforms compared to other models, such as XGBoost, thus a general architecture that accommodates gradient-boosted models, as opposed to the previously realized accelerator \cite{pedretti_tree-based_2021} for RF-only is vital.

Fig. \ref{fig:dse}(a) shows as an example the Root Mean Square Error (RMSE) as a function of the model size expressed as the size of an analog CAM-based hardware, namely $N_{trees}\times2^D$, with $D$ maximum depth and $N_{trees}$ number of trees in the case of Rossmann stores sales dataset\cite{rossmann_store_sales}.
We repeated the same experiments for all datasets (Fig. \ref{supp-si:fig:dse} and \ref{supp-si:fig:dse2}), picking as design choices for our architecture $N_{trees}=4096$ and $D=8$, corresponding to a total of 4096 \textit{cores} of with $2^8=256$ \textit{words} each.
The dashed line corresponds to our selected maximum model size, and thus hardware size and light blue area correspond to the model that we can map to the hardware.
It is possible to see how the accuracy saturates at a certain model size regardless of the configuration chosen (i.e. 4 or 8-bit, RF-only, or any tree-based model).

Results show that RF models are insufficient to reach good precision, and 8-bit inference can increase accuracy up to 2\% for classification and a $\sim$20\% reduction on RMSE for regression.
Fig. \ref{fig:dse}(b) shows the accuracy and RMSE for multiple datasets assuming constraint-free hyperparameters optimization (GPU), limited to our design choice with 8-bit (X-TIME 8-bit) and 4-bit (X-TIME 4-bit), and using only RF models (RF Only) demonstrating that (1) 8-bit can match the constraint-free design, (2) 4-bit is not enough for some datasets and (3) RF is underperforming significantly the best models for some datasets.
\ul{A printout of the data can be also found in Table }~\ref{supp-si:tab:accuracy}.

The final model and parameter selections for each dataset are reported in Table \ref{table:datasets}.
\ul{Given that the hardware cost scales aggressively with $D$ but linearly with $N_{trees}$ (Fig. }\ref{supp-si:fig:cost}\ul{), and the accuracy saturates for $D>8$}, we designed X-TIME for being able to perform inference of models up to $N_{trees} = 4096$, $D=8$, and $N_{bit} = 8$.
\ul{If more trees are needed, multiple X-TIME chips can be operated in parallel, similarly to how multiple cores are operated in parallel}.

\section{X-TIME Architecture}
\begin{figure}
\centering
\includegraphics[width=0.99\linewidth]{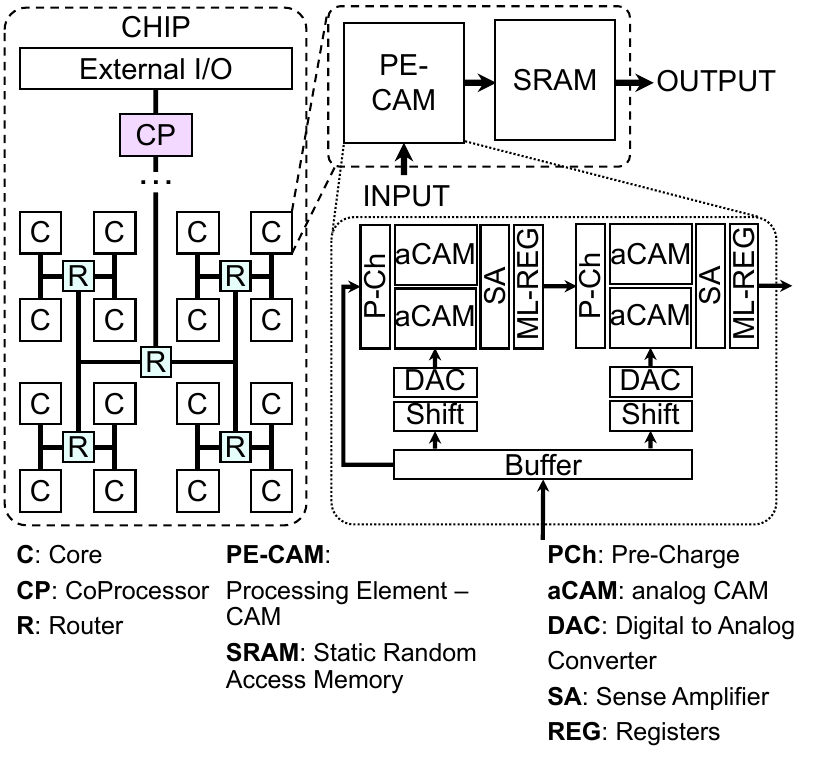}
\caption{Architecture hierarchy} 
\label{fig4}
\end{figure}
Fig.~\ref{fig4} shows the internal architecture of an X-TIME single-chip accelerator. 
The accelerator is composed of a group of cores connected by an H-tree NoC topology converging to a co-processor (CP).
Each core contains an analog CAM array acting as a processing element (PE-CAM) and an SRAM memory. 
To map models larger than a single array, each PE-CAM has \textit{stacked} arrays (extended row-wise) and \textit{queued} arrays (extended column-wise) to handle more and larger words, respectively. 
Stacked arrays share the same peripherals, namely the Pre-Charger (P-Ch), the Digital Analog Converter (DAC), the Sense Amplifier (SA), and the Match-Line Register (ML-REG). 
The connection of queued arrays ensures that only previously matched lines are charged by making the ML-REG of array $i$ feed the P-Ch of the array $i+1$.

\subsection{Inference}
The X-TIME Compiler (X-TIME-C) maps the trained trees to the cores according to Fig.~\ref{fig3}.
X-TIME-C is based on Treelite \cite{treelite}, a universal model exchange and serialization format for tree-based models.
Treelite outputs a C-like code with conditional statements representing each tree in the ensemble as an intermediate representation for further compilation to the hardware.
X-TIME-C converts this code for analog CAM usage, obtaining a table of size $L\times(2N_{feat}+3)$ with each row storing the lower/upper bound for each feature, the leaf value, class ID/label (in the case of multi-class classification) and tree ID, representing the model ensemble for our architecture. 
X-TIME-C assigns trees to cores via round-robin, directly mapping the first $L\times(2N_{feat})$ columns of a tree to the lower and upper bound of the analog CAM circuit, and the third-last column to the SRAM. 
Class and tree ID are uniquely represented in the core address.
Fig. \ref{supp-si:fig:compiler} shows a schematic visualization of X-TIME-C compilation flow.

The largest ensemble that can be mapped to X-TIME is constrained by the maximum number of leaves $N_{leaves,max}$ in all trees and the number of available cores $N_C$. 

During inference, a feature vector is sent through the network to a core and searched in the PE-CAM. 
Each queued analog CAM block receives a different portion of the feature vector which is converted into analog voltage values by a DAC. 
A logical AND operation is inherently performed between the queued arrays, similar to ML segmentation used in modern TCAMs to maintain high throughput where MLs are divided into sections of shorter words with an AND between individual ML results. 
Connecting multiple TCAM arrays with common MLs is logically equivalent to a larger TCAM, with the bit-wise AND operation between TCAM cells on the same ML now extended between arrays. 
In our case, the analog CAM stores a branch split into multiple analog CAM arrays and reconstructed using the logical AND, such that only if all branch nodes arranged in multiple arrays are matched will a match be returned on the corresponding MAL.

The final output of the PE-CAM is stored in a register and used to address the SRAM, storing the leaf value of each branch.

Leaf values coming from multiple cores are summed with a programmable adder tree to perform the reduction operation.
Finally, the resulting leaf value is propagated to the CP which performs a \textit{global} reduction and computes the model prediction. 
The latter is usually a simple operation, such as a majority voting or a threshold comparison.
In the case of a model larger than what can be mapped in a single X-TIME chip, we foresee the possibility of multiple CPs in multiple chips orchestrating a reduction similar to the adder tree in the chip network.

\ul{X-TIME is an inference accelerator, but models in production might be retrained and updated, due to data and/or concept drift, up to every 100 days}~\cite{vela2022temporal}. 
\ul{While ReRAM endurance might be limited to 10$^5$-10$^8$ cycles depending on the technology}~\cite{mannocci2023memory}\ul{ given the large retraining cycle, the lifetime of the accelerator is significantly larger than the lifetime of typical systems, making ReRAM endurance not a strong requirement.} 

\subsection{Analog CAM with improved precision}\label{section:precision}
\begin{figure}
\centering
\includegraphics[width=0.98\linewidth]{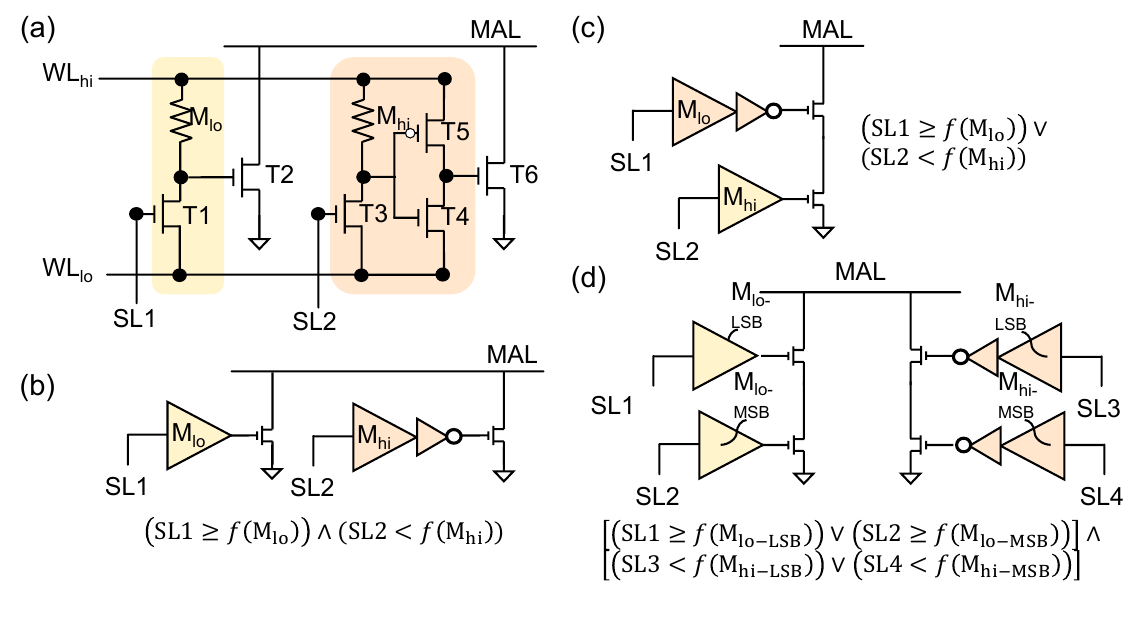}
\caption{(a) Circuit schematic and (b) block diagram of a \textit{AND-type} 4-bit 6T-2M analog CAM~\cite{li_analog_cam_2020}. (c) Block diagram of a 4-bit OR-type analog CAM. (d) Block diagram of an 8-bit analog CAM.}
\label{fig5}
\end{figure}

As a result of experiments in section \ref{sec:training} we found out that typically 8 bits precision, corresponding to 256 bins for each feature, is required for matching floating point (FP) accuracy inference of tree-based ML models.

While analog CAM circuits offer the possibility of mapping arbitrary ranges, RRAM device reliability poses a challenge in supporting more than 16 levels or 4 bits~\cite{pedretti_tree-based_2021}. 
In the case of crossbar arrays mapping algebraic matrices, such as a neural network layer, multiple techniques have been proposed for overcoming limited device precision.
For example, with bit slicing~\cite{shafiee_isaac_2016} by using two $M$ bits RRAMs, $N=2^{2M}$ bits can be represented by connecting two devices on the same row (Fig. \ref{supp-si:fig:slicing}(a)). 
In crossbar arrays, it is also straightforward to arbitrarily adjust the input precision by applying binary pulses and performing SHIFT and ADD operations at the output~\cite{shafiee_isaac_2016}. 

Unfortunately, such techniques do not apply to analog CAM arrays. 
The analog CAM encodes a \textit{unary} representation and by connecting two analog CAM cells on the same MAL to represent one macro-cell, it is only possible to double the number of levels (i.e. $2^{N_{bits}}$) rather than the number of bits. 
Thus given RRAMs with $M = 4$ bits, \textbf{16} analog CAM cells are needed for representing $N = 8$ bits, leading to an exponential overhead (Fig. \ref{supp-si:fig:slicing}(b)).
Here, we propose a novel solution to increase the precision with the same base analog CAM cell circuit, which requires 2 analog CAM cells and 2 clock cycles for performing 8-bit search operation with $M = 4$ bits RRAMs.

To demonstrate, let us first take the case where the lower bound threshold $T_{L}$ is needed to be mapped and operated with $N=8$ bit of precision but we only have $M=4$ bits of precision in our analog CAM and RRAMs devices. 
The lower bound threshold can be sub-divided into its most significant bit representation $T_{LMSB}$ and least significant bit representation $T_{LLSB}$, each one made of $M$ bits and written as $T_{L}=2^M T_{LMSB}+T_{LLSB}=16 T_{LMSB}+T_{LLSB}$.
Each analog CAM cell computes the conditional operation $\textrm{MAL} = q\geq T_L$, where $q$ is the input query applied to the DL. 
We similarly divide the input $q$ into most and least significant representations $q_{MSB}$ and $q_{LSB}$, such that $q=2^M q_{MSB}+q_{LSB}=16 q_{MSB}+q_{LSB}$ and we can obtain the result of $\textrm{MAL}$ by computing
\begin{equation}\label{eq:double1}
\begin{split}
    [(q_{MSB}\geq T_{LMSB})\wedge (q_{LSB}\geq T_{LLSB})]\\
    \vee (q_{MSB}\geq T_{LMSB}+1)
\end{split}
\end{equation}
or alternatively, rearranging the terms:
\begin{equation}\label{eq:double2}
\begin{split}
    [(q_{MSB}\geq T_{LMSB}+1)\vee (q_{LSB}\geq T_{LLSB})]\\
    \wedge (q_{MSB}\geq T_{LMSB}).
\end{split}
\end{equation}
Note that Eq. \ref{eq:double2} is more CAM-friendly since it is a conjunction of two terms that can be appended to the same MAL. 
The same calculation derives the upper bound $T_H$ conditions and with an AND operation between this and equation (\ref{eq:double2}), the full computation $\textrm{MAL} = T_L\leq q < T_H$ results in:
\begin{equation}\label{eq:double3}
\begin{split}
    [(q_{MSB}\geq T_{LMSB}+1)\vee (q_{LSB}\geq T_{LLSB})]\\
    \wedge (q_{MSB}\geq T_{LMSB})\\
    \wedge [(q_{MSB}< T_{HMSB})\vee (q_{LSB}< T_{HLSB})]\\
    \wedge (q_{MSB}< T_{HMSB}+1).
\end{split}
\end{equation}
Analog CAMs cells arranged on the same \textrm{MAL} perform a bit-wise AND ($\wedge$) operation between them, thus a circuit addition is needed for performing the OR ($\vee$) operation. 
We start from the \textit{conventional} analog CAM cell shown in Fig. \ref{fig5} (a).
The lower bound (yellow, T1, $M_lo$) is encoded in a 1-transistor-1-resistor structure.
If the input is high enough, it turns on T1, lowering the gate node of the pulldown T2 which stays off, resulting in a match.
If the input voltage is low or alternative $M_lo$ conductance is high, T2 is turned on.
Thus the lower bound checks if $SL_1>f(M_lo)$.
A similar discussion can be applied to the higher bound (orange, T3, $M_hi$, T4, T5), in which an inverter (T4,T5) is added between the input voltage divider (T3,$M_hi$) and the pulldown (T6) to represent the higher bound and checking if $SL_2\leq(M_hi)$.
Fig. \ref{fig5} (b) shows a more compact circuit schematic for the analog CAM in Fig. \ref{fig5}(a), with comparators driving the pulldown representing the 1T-1M and 1T-1M-1INV structures.
We refer to this circuit as \textit{AND-type}, given the AND operation between the lower and upper bound.
Similarly, an \textit{OR-type} circuit can be realized (Fig. \ref{fig5} (c)) \cite{pedretti_differentiable_correct}, effectively performing an \textit{out of range} matching.
Finally, by combining the two structures, it is possible to realize the general circuit of Fig. \ref{fig5} (d) which can implement Eq. \ref{eq:double3}, by applying opportunely inputs to the comparator (see Supplementary Information  \ref{supp-si:section:8bit-acam})
Note that separate inputs for each cell are required for the programming operation, thus no extra wiring to the cell is required.

\section{Benchmark Results}
\subsection{Comparison with state-of-the-art}
\begin{figure*}
\centering
\includegraphics[width=\linewidth]{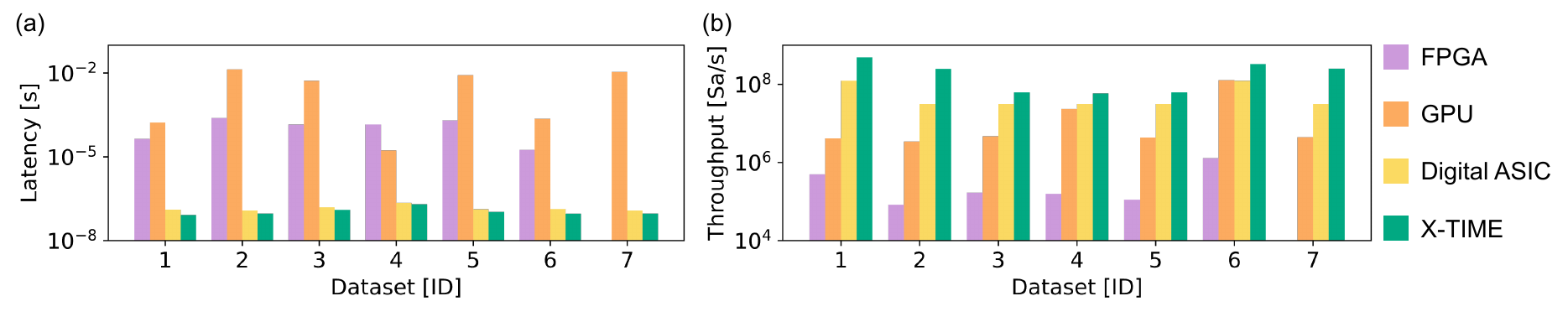}
\caption{Comparison of (a) latency and (b) throughput of this work, GPU, FPGA, and fully digital ASIC accelerator from Fig. \ref{supp-si:fig:digital_accelerator}.} 
\label{fig10}
\end{figure*}
We benchmarked X-TIME using our open-sourced cycle-approximate simulator\footnote{https://github.com/HewlettPackard/X-TIME} on a wide range of models and dataset types (Table~\ref{table:datasets}), and compared latency and throughput (see Fig.~\ref{fig10}(a,b)) to GPU, \ul{specifically NVIDIA V100}, FPGA~\cite{gajjar_faxid_2022} and a digital accelerator based on small SRAMs and similar to Booster~\cite{he_booster_2022} (Fig. \ref{supp-si:fig:digital_accelerator}). 
More details on the methodology can be found in Supplementary Information \ref{supp-si:section:soa}, \ref{supp-si:section:methods} \ul{and Table}~\ref{supp-si:tab:benchmark}.
When possible, we applied input batching and replicated trees to increase throughput (Fig. \ref{supp-si:fig:batching}).
As shown, our architecture produces drastically reduced latency, frequently to $\sim$100ns compared to GPU at 10 $\mu$s to ms, while also demonstrating significantly improved throughput by 10-120$\times$ across most datasets. 
Our architecture shines for large binary classification or regression models with a peak improvement in latency by $9740\times$ at 119$\times$ higher throughput for the Churn modeling dataset inference \cite{churn_modelling} compared to GPU.
As tree search is parallel in each analog CAM array of each core, as well as the in-network reduction, latency, and throughput are constant with $N_{trees}$ and $N_{leaves}$ with a delay solely dependant on $N_{feat}$ and $N_{classes}$. 
In contrast, on GPU there is a linear dependence on $N_{trees}$ and $D$, which limits the performance for large models. 
In the FPGA implementation, namely FAXID, there is a linear dependence on $N_{trees}$.
The SRAM-based accelerator has a linear dependence on $D$ (since multiple memory accesses are required) and $N_{trees}$ when more than a tree needs to be mapped in a core. 
\begin{figure}
\centering
\includegraphics[width=0.8\linewidth]{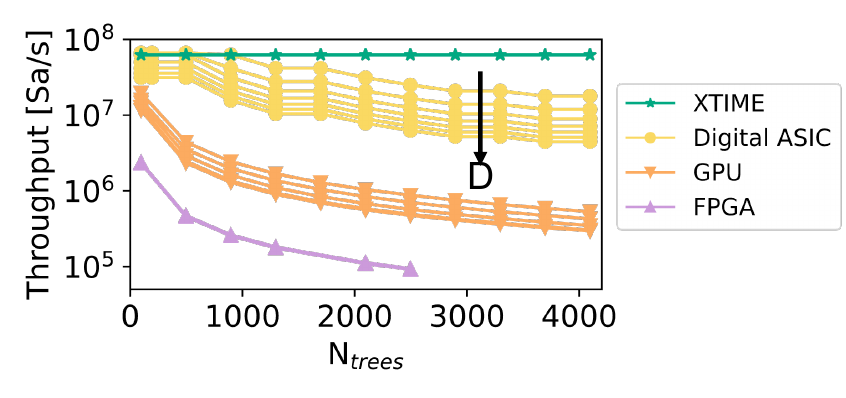}
\caption{Throughput as a function of number of trees $N_{trees}$ and depth $D$ for various accelerators. Note that in the case of X-TIME and the FPGA implementation, there is no dependence on the depth, removing the load imbalance issue, while only X-TIME shows no dependence on the number of trees.} 
\label{fig11}
\end{figure}
As an example, we studied the throughput as a function of $N_{trees}$ and $D$ for GPU, FPGA, SRAM-based accelerator, and X-TIME as shown in Fig.~\ref{fig11}, for a multiclass classification dataset \cite{blackard1999comparative}, demonstrating the predicted trend. 
We note that the SRAM-based accelerator still suffers from load imbalance between trees, similar to GPU, in which trees with different depths have different inference times, which results in the need for synchronization before the reduction operation.
As an energy comparison with GPU, we considered a worst-case scenario.
Fig. \ref{supp-si:fig:energy} shows the energy saving in comparison by considering the peak power consumption (19 W) and idle power consumption (25 W) in the case of X-TIME and GPU, respectively.
Even by assuming a full chip utilization for X-TIME and no energy consumption for the tree inference on GPU, X-TIME is up to 157$\times$ more energy efficient. 

\subsection{Sensitivity to noise}\label{noise}
\begin{figure}
\centering
\includegraphics[width=0.99\linewidth]{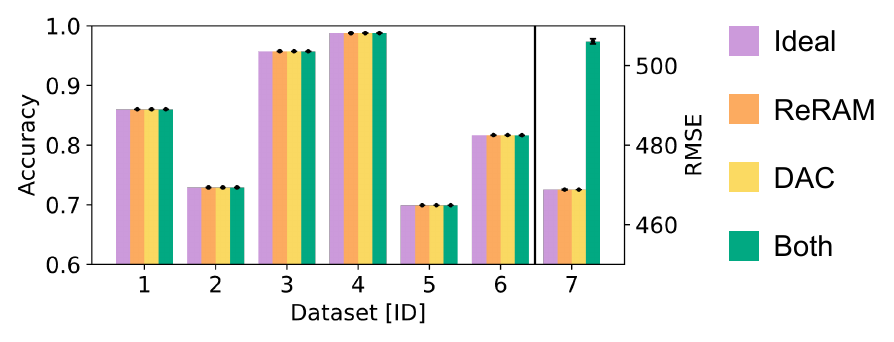}
\caption{Accuracy and RMSE for different noise configurations} 
\label{fig:noise}
\end{figure}
Analog hardware is subject to noise and non-idealities.
However, tree-based ML ensembles are resilient to variation thanks to the \textit{averaging} mechanism happening between trees of multiple boosting rounds.
Fig. \ref{fig:noise} shows the average accuracy of the models for different noise profiles, namely no noise (ideal), only ReRAM or DAC noise, and both noise sources, over 100 inference experiments.
\ul{A printout of the data can also be found in Table }~\ref{supp-si:tab:noise}.
We characterized the noise of TaOx-based ReRAM devices at various conductance values ranging from $<0.1$ nS to 500 $\mu$S \cite{sheng2019low}, by reading them 1000 times and extracting mean and standard deviation.
We measured a maximum relative standard deviation of noise of $\sigma_G/G=0.1$.
While DAC designs can reach higher precision, only 4-bits are needed for X-TIME inference.
Nevertheless, we considered a pessimistic standard deviation of noise for the DAC $\sigma_{DAC}=50mV$.
Results indicate no perturbation in the accuracy and RMSE when either ReRAM or DAC noise is applied, thanks to our hardware-aware training and inherent averaging mechanisms of ensemble models.
In the case of regression and with both noise sources applied there is an RMSE perturbation, but still reduced compared to the 4-bit operation (Fig. \ref{fig:dse}).

Intuitively, with the thresholding mechanism of Analog CAM, it is possible to program conductances with a separation such that the probability of exceeding the threshold is relatively low, leading to good noise resistance performance.
\section{Conclusion}
In this work, we presented a novel architecture for accelerated tree-based ML inference with in-memory computing based on a novel increased precision analog non-volatile CAMs. Analog CAMs enable novel applications of in-memory computing beyond matrix multiplies. Moreover, the required peripherals are significantly reduced compared to crossbar-based computing primitives. Our approach maps decision trees to analog CAM ranges, which directly compute the matched results in parallel, given an input feature vector. Leaf values are stored in a local SRAM and accumulated at the core level when multiple trees are mapped in the same core. Outside the core, an H-tree NoC connects 4096 cores and a co-processor in a single chip. An in-network computing structure accumulates leaf values as data flows to limit data movement. The on-chip network is re-configurable to map different models (regression, multi-class classification, etc) and inference techniques, such as input batching. 
An HW-aware training routine ensures to gets state of the art accuracy by leveraging quantization and optimizing the number of trees and maximum depth.
Our results show significant improvement in latency -- by almost four orders of magnitude, compared with an NVIDIA V100 GPU, at a throughput up to $\sim120\times$ higher across a wide range of dataset types and model topologies. 
Further, the worst case scenario showed up to $150\times$ improvement in energy efficiency compared to GPUs with our approach.

\bibliographystyle{IEEEtran}
\bibliography{main}
\clearpage

\end{document}


\receiveddate{XX Month, XXXX}
\reviseddate{XX Month, XXXX}
\accepteddate{XX Month, XXXX}
\publisheddate{XX Month, XXXX}
\currentdate{XX Month, XXXX}
\doiinfo{JXCDC.2023.3238030}

%

\title{\textcolor{ojcolor2}{Supplementary Information for X-TIME: accelerating large tree ensembles inference for tabular data with analog CAMs}}

\author{GIACOMO PEDRETTI, MEMBER, IEEE, 
        JOHN MOON, 
        PEDRO BRUEL, 
        SERGEY SEREBRYAKOV,
        RON M. ROTH, FELLOW, IEEE,
        LUCA BUONANNO,
        ARCHIT GAJJAR,
        LEI ZHAO,
        TOBIAS ZIEGLER,
        CONG XU, 
        MARTIN FOLTIN, 
        PAOLO FARABOSCHI, FELLOW, IEEE,
        JIM IGNOWSKI, SENIOR MEMBER, IEEE,
        CATHERINE E. GRAVES}

\affil{Artificial Intelligence Research Lab (AIRL), Hewlett Packard Labs, Milpitas, CA 95035 USA}
\affil{Artificial Intelligence Research Lab (AIRL), Hewlett Packard Labs, Fort Collins, CO 80528 USA}
\corresp{CORRESPONDING AUTHOR: Giacomo Pedretti (e-mail: giacomo.pedretti@hpe.com).}
\authornote{This work was funded by Army Research Office under Agreement W911NF2110355.}
\maketitle

\section{Tree-based ML model inference on GPU}\label{si:section:gpu_inference}
Recent NVIDIA libraries implementations have shown order of magnitude improvements compared to CPU and previous GPU implementations ~\cite{raschka2020machine}. 
However, three different factors still limit GPU throughput and latency~\cite{xie_tahoe_2021}, in particular for tree ensembles. 
First, a high number of uncoalesced memory accesses, due to the low ratio between requested and fetched data, causes low memory efficiency. 
Trees are usually stored in memory with nodes from different trees interleaved in breadth-first order, resulting in coalesced memory accesses close to the root, where children nodes are more likely to be part of the same memory transaction of the parent node, but uncoalesced memory accesses further from the root. 
Thus, with increasing tree depth the proportion of coalesced memory accesses decreases and memory system efficiency also decreases.
Second, synchronization between threads and the consequent load imbalance due to waiting for the deepest trees limits latency. 
A common implementation scheme on GPU assigns each tree in the ensemble to a different thread by round-robin. 
At the end of each thread block, a reduction operation is then performed. 
However, a given ensemble may contain \textit{short} and \textit{tall} trees, with the latter having substantially larger latency. 
Thus, each thread block must wait until the slowest (tallest) tree has finished processing, increasing overall latency. 
Finally, with a larger number of trees $N_{trees}$ in the ensemble, the increasing number of thread blocks and the global reduction performed between thread blocks becomes a significant overhead. 
Multiple techniques have been proposed to increase the throughput of tree ensemble inference~\cite{raschka2020machine,xie_tahoe_2021}, mainly by re-organizing the memory - for example, for sparse trees and scheduling of threads - but with limited impact on latency. 

\section{Input signal conditioning for 8-bit analog CAM operation}\label{si:section:8bit-acam}

\begin{table}[h!]
  \center
  \caption{Input to apply for doubling precision with time shifting}\label{table:shifting}
  \begin{tabular}{lrrrrrrrrrrrr}
  \toprule
    \textbf{Input} & \textbf{Cycle 1} & \textbf{Cycle 2}\\
\midrule
    $q_{HLSB}$ & $q_{LSB}$ & $GND$ \\
    $q_{LLSB}$ & $q_{LSB}$ & $VDD$ \\
    $q_{HMSB}$ & $q_{MSB}$ & $q_{MSB}-1$ \\
    $q_{LMSB}$ & $q_{MSB}-1$ & $q_{MSB}$ \\
\bottomrule
  \end{tabular}
\end{table}

Table~\ref{table:shifting} shows the required inputs for each DL analog bus wire to compute the 8-bit analog CAM equation in a 2-step search operation, namely

\begin{equation}\label{si:eq:double3}
\begin{split}
    [(q_{MSB}\geq T_{LMSB}+1)\vee (q_{LSB}\geq T_{LLSB})]\\
    \wedge (q_{MSB}\geq T_{LMSB})\\
    \wedge [(q_{MSB}< T_{HMSB})\vee (q_{LSB}< T_{HLSB})]\\
    \wedge (q_{MSB}< T_{HMSB}+1).
\end{split}
\end{equation}

With MAL pre-charged high, the first cycle evaluates the first bracket of equation (\ref{si:eq:double3}) for the lower and upper bound (the OR operation). 
Without pre-charging MAL again, the second cycle evaluates the second term of each bound by inputting \textit{always care} (i.e. always mismatch) values to the LSB sub-cell. 
Therefore, an AND operation between the outputs of cycle 1 and cycle 2 is computed on MAL following cycle 2, similar to a 2-step search approach previously demonstrated for increasing bit density in TCAMs~\cite{lin_74_2016}.
To the best of our knowledge, this is an entirely new approach and the proposed circuit operation is the most compact way of computing equation (\ref{si:eq:double3}) with an analog CAM, and thus for doubling the bit precision. 

\begin{figure*}[h!]
\centering
\includegraphics[width=0.9\linewidth]{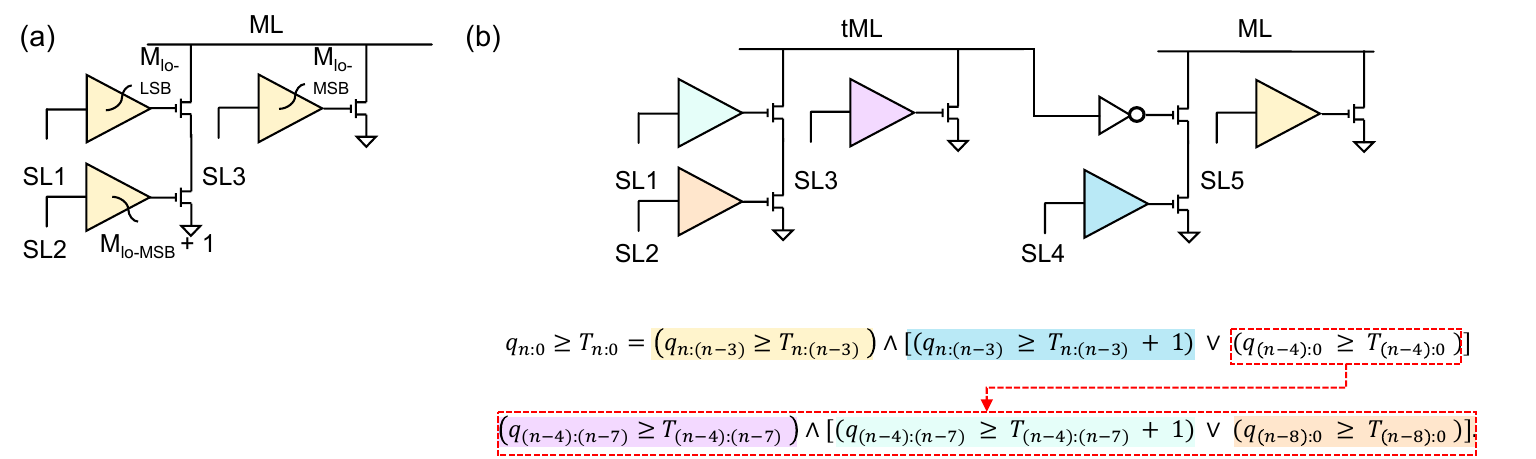}
\caption{(a) Single-cycle doubled precision cell and (b) its extension to arbitrary precision.} 
\label{si:fig:recursive}
\end{figure*}

Other circuit designs may implement the same function in one clock cycle, these would have a significantly larger area. 
For example, one could separate cells performing the OR operation by connecting the discharge transistors in series rather than in parallel \ul{as shown in Fig.}~\ref{si:fig:recursive}a \ul{for the lower bound case}.
This approach would also allow for selecting the precision of each feature to be either 4 or 8-bit, depending on the required precision. 
\ul{Moreover, the circuit can be extended to arbitrary precision.
Assuming for example a required precision of $N=12$ bits and ReRAM with $M = 4$ bits, the first comparison in Eq.}~\ref{si:eq:double3}\ul{ can be expanded recursively with a similar approach}.
\ul{Fig.} ~\ref{si:fig:recursive}b \ul{shows an example (lower bound only) of an Analog CAM for $N = 12$ bits computation using the proposed approach, with the inset showing the recursive equation}.
\ul{More in general, given $(k-1)M\leq N \leq kM$, with $k$ a parameter, $1+3(k-1)$ 6T2M Analog CAM cell are needed for representing the higher precision comparison.}

In any case, given that analog CAM search latency is forecast to 100 ps~\cite{li_analog_cam_2020}, we found that performing two searches rather than one would not unduly increase the overall latency.
Thus, we adopt the circuit formulation and operation in the main text, which doubles the bit precision while doubling the area and analog CAM search latency.
\section{Evaluation Methodology}\label{evaluation}\label{si:section:methods}
We developed a functional and cycle-detailed simulator\footnote{https://github.com/HewlettPackard/X-TIME} to evaluate latency,
throughput, energy efficiency, area, and accuracy of the in-memory engine for tree-based model inference. 
The architecture simulator was developed using the Structural Simulation Toolkit (SST)~\cite{rodrigues_structural_2011} and estimates the performance on the benchmark datasets detailed above. 
SST, which is a parallel discrete-event simulator framework, can integrate modular components in a unified interface and simulate large-scale systems with MPI. 
It also provides interfaces for simulation components and components that perform actual functions such as processors, memory, and networks. For our system, the user-defined components emulate the cycle-level function of each X-TIME architecture, and data flow through the links connecting the components with specified delay. 
The cycle-level architecture simulation consisting of an H-tree NoC with 4096 cores and 1365 routers verified the error-free functional operation and the pipelined architecture without hazards, while also yielding system performance estimates.
Fig.~\ref{fig8} summarizes the power and area for the proposed architecture (full details in Supplementary Table \ref{si:table:powerarea}). 
We consider the previously presented 16nm analog CAM model ~\cite{li_analog_cam_2020,pedretti_tree-based_2021}, which included the analog CAM array, DAC, SA, and P-Ch area, latency, and power consumption. 
We considered ReRAM conductances from $1\mu S$ to $100\mu S$ to map the required 16 levels for each device. 
The analog CAM model also includes parasitics that limit the CAM pre-charge and discharge time and the DAC operational frequency~\cite{li_analog_cam_2020}.
Power and area values for register and logic are based on TSMC 16nm PDK. 
Notably, the analog CAM arrays dominate area and power consumption, with peripheral components contributing negligibly. 
\ul{Note that we did not consider the communication time and energy in and out of the chip. For a fair comparison, we did not consider it also for other accelerators in the benchmark, such as the GPU.}

\section{Comparison with GPU, FPGA and SRAM based accelerators}\label{si:section:soa}

To demonstrate the analog CAM performance capabilities, we compare our results to three alternative accelerator approaches: GPU, FPGA, and digital SRAM. As our first performance baseline, we profiled the inference step of the tree models and corresponding datasets on an NVIDIA V100 GPU. 
We measured each GPU kernel execution time with 5 repetitions to process data batches of increasing size, up to a saturation point, where throughput improvements were no longer observed. 
To estimate GPU upper bound performance, we isolated the kernel execution time of each model using NVIDIA's \textit{nvprof} tool and did not include the memory transfer times between host (CPU) to device (GPU) or internal GPU transfers in our measured times. 
When possible, we profiled models using CUDA kernels from the NVIDIA RAPIDS Forest Inference Library (FIL).

Moreover, we compared X-TIME with an FPGA baseline FAXID \cite{gajjar_faxid_2022}. 
FAXID is a custom hardware accelerator for low-latency ransomware detection using high-level synthesis (HLS) using binary classification models. As FAXID only supports binary classification, for our comparison we used a 1-vs-all approach to benchmark latency and throughput where we selected one of the classes as class 0 and set the remaining classes to class 1. 
In FAXID the maximum depth was restricted to 6, as their ransomware detection accuracy did not increase with deeper trees. To keep the baseline results faithful to the FAXID HLS design, we optimized the hyperparameters for smaller depth and a larger number of trees on FAXID to match accuracy with the other approaches.

Finally, we compare X-TIME to a custom digital accelerator baseline using SRAM to isolate the performance benefits from the analog CAM specifically. 
This SRAM baseline uses a distributed architecture with each tree mapped to a small SRAM accessed serially by a floating point unit \cite{he_booster_2022}, as shown in Fig. \ref{si:fig:digital_accelerator}. 
Each SRAM word stores the node value and feature, the left and right children addresses, or the corresponding logit or class in the case of a leaf node.
We considered the X-TIME network architecture and changed just the core operation (analog CAM or digital SRAM) to simulate the effect on throughput, of multiple accesses required for each tree. 

One might consider a digital TCAM to be another meaningful comparison. However, as previously demonstrated \cite{roth2023implementation}, using a traditional digital TCAM for performing on-access comparison would result in large overheads. In detail, $2^8$ TCAM columns would be needed to represent each feature with 8-bit precision, and thus tree inference of 129 features requires 33024 TCAM columns. Considering an 80-column TCAM macro has an access time of $\sim500 ps$  \cite{tsukamoto2015mbit}, inference with a TCAM would take $500\cdot33024/80 ps \approx20 us$ per tree, significantly larger than the few cycles analog CAM latency.

\clearpage

\section{Supplementary Figures}

\begin{figure}[h!]
\centering
\includegraphics[width=0.65\linewidth]{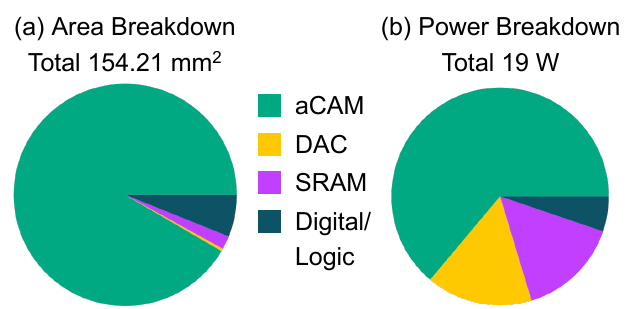}
\caption{Area (a) and peak power (b) breakdown} 
\label{fig8}
\end{figure}

\begin{figure*}[h!]
\centering
\includegraphics[width=0.9\linewidth]{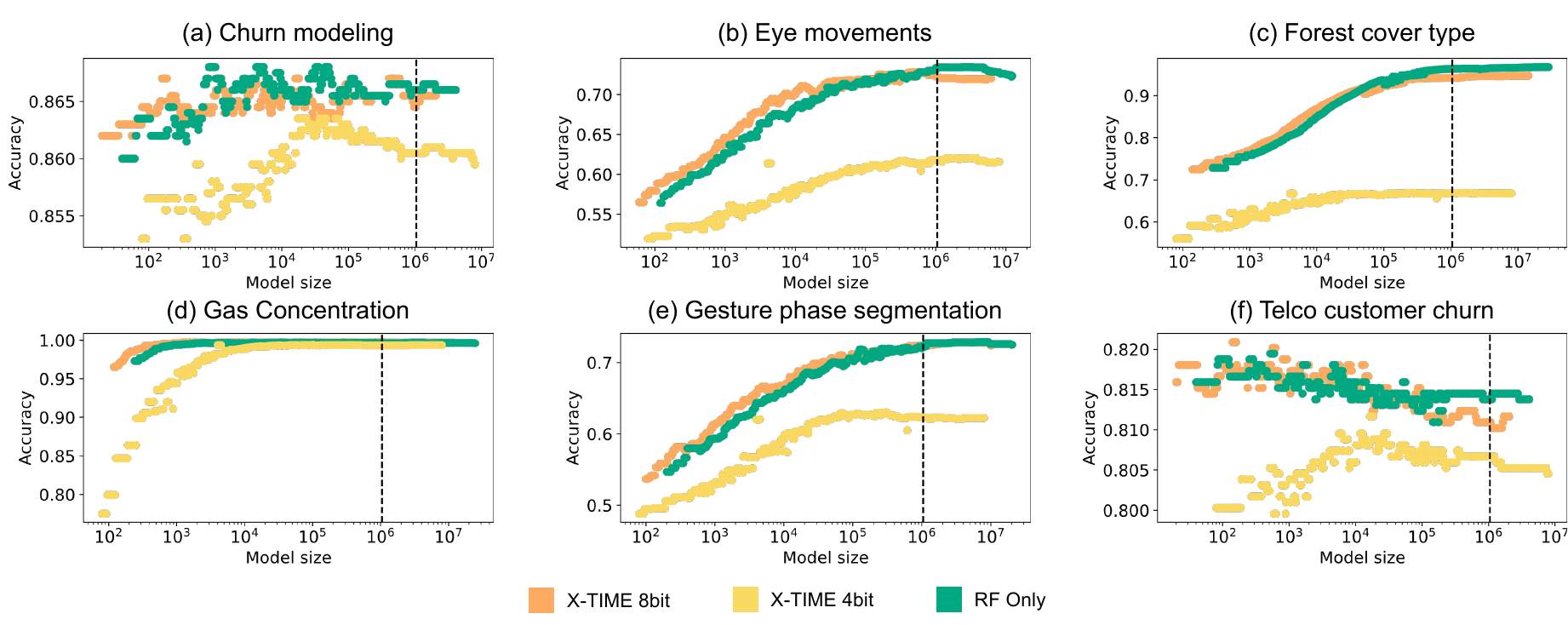}
\caption{Accuracy as a function of model size for multiple datasets and training methodologies} 
\label{si:fig:dse}
\end{figure*}

\begin{figure*}[h!]
\centering
\includegraphics[width=0.9\linewidth]{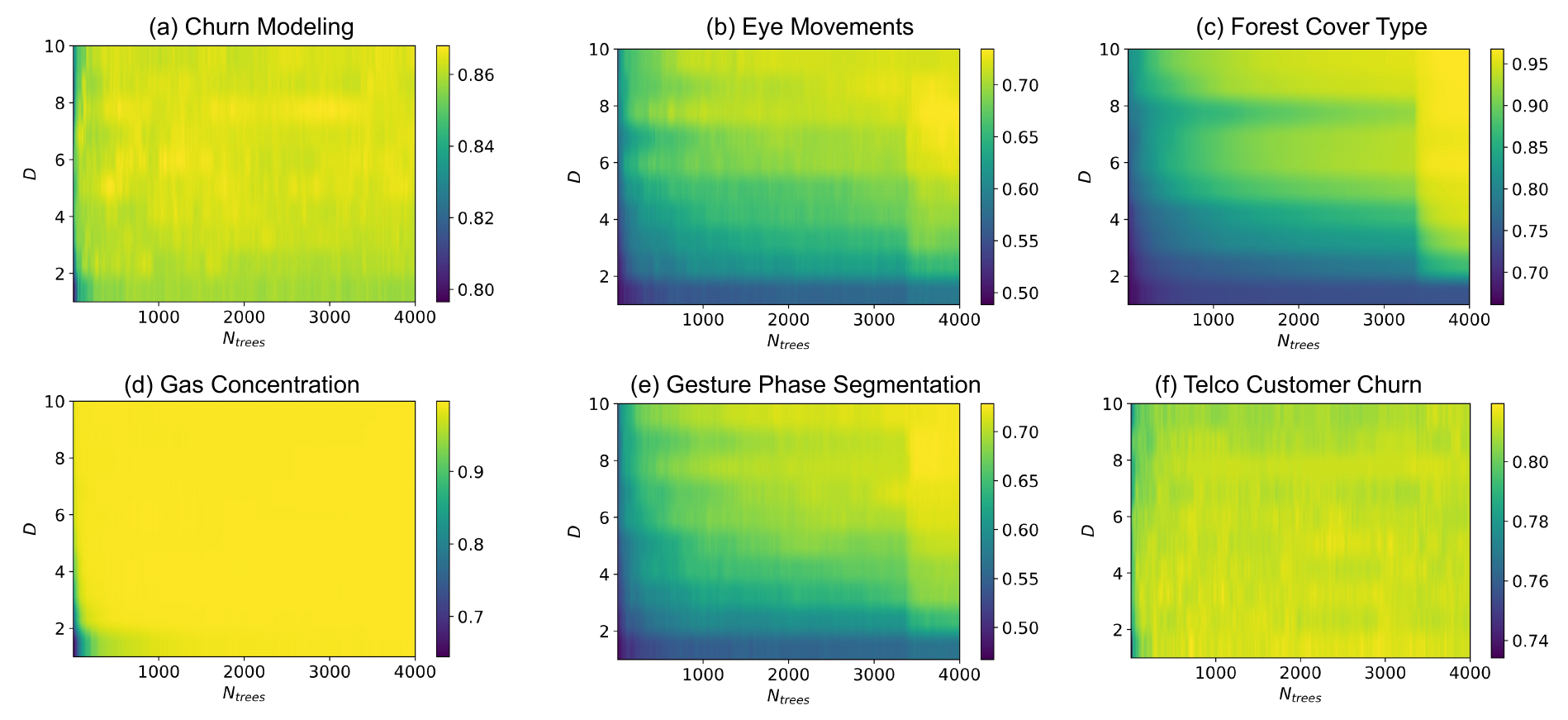}
\caption{Accuracy training with XGBoost as a function of depth and number of trees for multiple datasets.} 
\label{si:fig:dse2}
\end{figure*}

\begin{figure}[h!]
\centering
\includegraphics[width=0.9\linewidth]{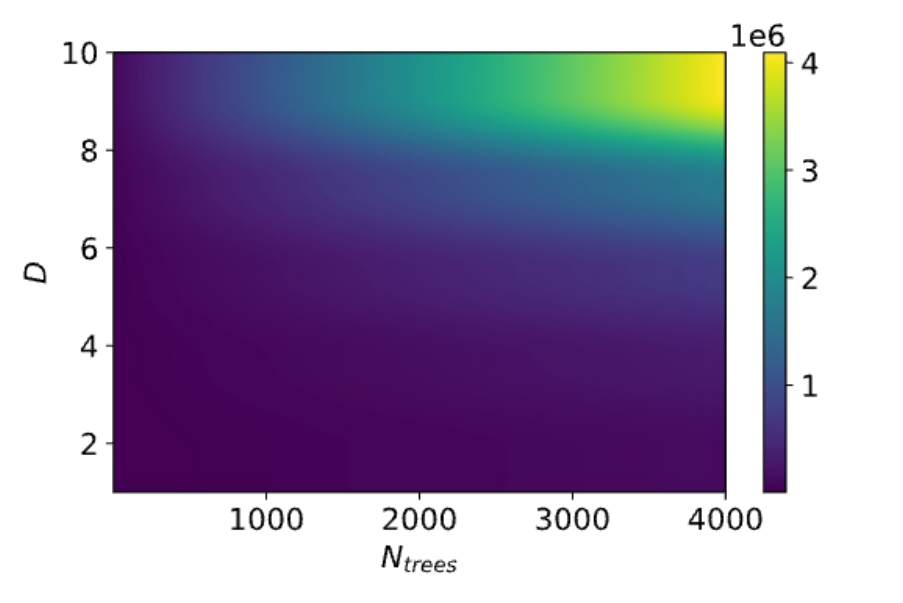}
\caption{Hardware cost in arbitrary unit as a function of depth and number of trees for multiple datasets.} 
\label{si:fig:cost}
\end{figure}

\begin{figure}[h!]
\centering
\includegraphics[width=0.8\linewidth]{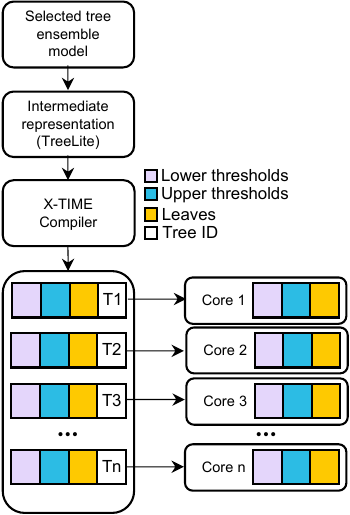}
\caption{Schematic representation of the X-TIME Compiler (X-TIME-C) flow.} 
\label{si:fig:compiler}
\end{figure}

\begin{figure}[h!]
\centering
\includegraphics[width=0.8\linewidth]{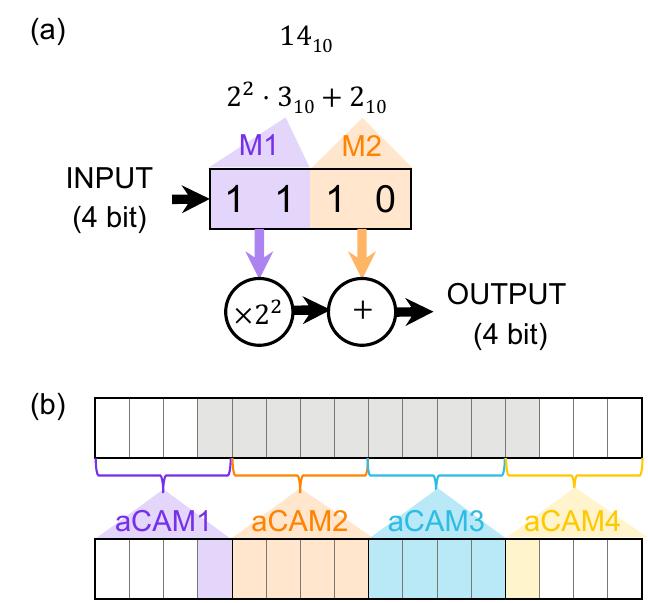}
\caption{(a) Bit slicing operation in crossbar arrays \cite{shafiee_isaac_2016} and (b) slicing operation in Analog CAM arrays resulting in exponential overhead.} 
\label{si:fig:slicing}
\end{figure}

\begin{figure}[h!]
\centering
\includegraphics[width=0.8\linewidth]{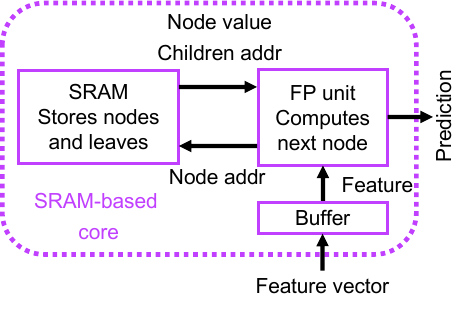}
\caption{Schematic of the SRAM-based accelerator core used as a digital baseline.} 
\label{si:fig:digital_accelerator}
\end{figure}

\begin{figure}[h!]
\centering
\includegraphics[width=0.8\linewidth]{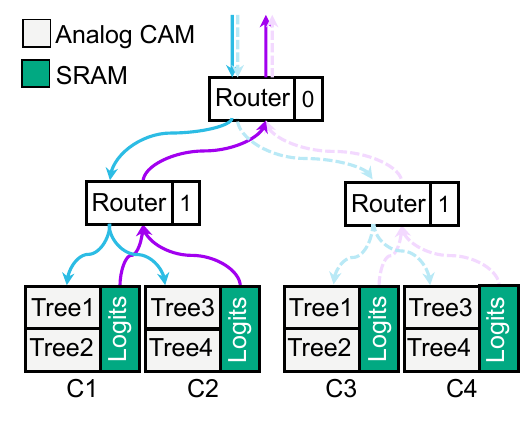}
\caption{Input batching operation for binary classification or regression. Trees are replicated in multiple cores, and multiple inputs (different colored arrows) are applied at the same time to increase the throughput and utilization.} 
\label{si:fig:batching}
\end{figure}

\begin{figure}[h!]
\centering
\includegraphics[width=0.9\linewidth]{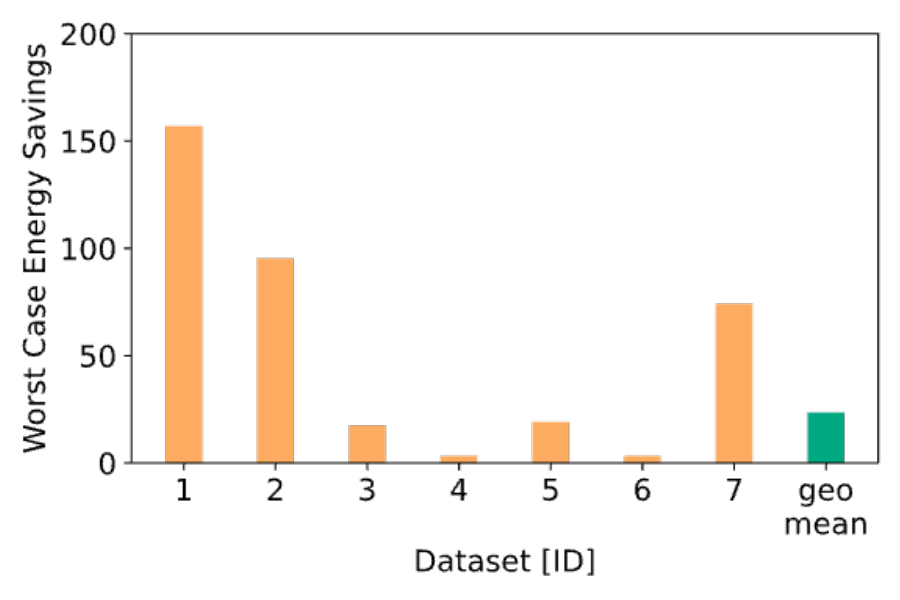}
\caption{Worst case scenario energy savings compared to GPU for multiple datasets (orange) and geometric mean (green)} 
\label{si:fig:energy}
\end{figure}

\clearpage

\section{Supplementary Tables}

\begin{table*}
    \centering
    \caption{Model performance for different hardware implementation, namely state of the art performance on GPU, 8-bit and 4-bit performance on X-TIME and limited to RF for previous work~\cite{pedretti_tree-based_2021}}
    \begin{tabular}{|c|c|c|c|c|c|}
    \hline
         \textbf{Dataset} &  \textbf{Metric} & \textbf{GPU} & \textbf{X-TIME 8bit} & \textbf{X-TIME 4bit}  & \textbf{RF Only}\\
    \hline
         Churn Modeling~\cite{churn_modelling} & Accuracy &	0.86 &	0.86 &	0.86 &	0.85 \\
    \hline
         Eye Movements~\cite{salojarvi2005inferring} & Accuracy & 0.75 & 0.73 & 0.73 & 0.50 \\
    \hline
         Forest cover type\cite{blackard1999comparative} & Accuracy & 0.96 &	0.96 &	0.940 &	0.73 \\
    \hline
         Gas concentration\cite{vergara_chemical_2012} & Accuracy &  0.99 &	0.99 &	0.81 &	0.81 \\
    \hline
         Gesture phase segmentation\cite{wagner2014gesture} & Accuracy &  0.7 &	0.7 &	0.69 &	0.34 \\
    \hline
         Telco customer churn\cite{telco_customer_churn}&  Accuracy &  0.81 &	0.81 &	0.81 &	0.81 \\
    \hline
         Rossmann stores sales\cite{rossmann_store_sales} & RMSE &  510.7	& 515.5 & 643.8 & 1257.6 \\
    \hline
    \end{tabular}
    \label{si:tab:accuracy}
\end{table*}

\begin{scriptsize}
\begin{table*}\label{si:table:powerarea}
  \centering
  \caption{Architecture parameters (* = simulated peak values)}
  \begin{tabular}{|l|l|l|l|l|}
  \hline
  \multicolumn{5}{c}{\textbf{Core}}\\
  \hline
  \thead{\textbf{Component}} & \thead{\textbf{Param}} & \thead{\textbf{Value}} & \thead{\textbf{Power [$\mu W$]}} & \thead{\textbf{Area [$\mu m^2$]}}\\
  \hline
  \makecell{Analog CAM\\array} & \makecell{\# rows\\\# cols\\Stacked\\Queued} & \makecell{128\\65\\2\\2} & \makecell{2881.23*} & \makecell{34504.7} \\
  \hline
  \makecell{DAC} & \makecell{\# channels\\\# bits} & \makecell{260\\4} & \makecell{769.04*} & \makecell{143} \\
  \hline
  \makecell{SA} & \makecell{\#} & \makecell{512} & \makecell{239.49*} & \makecell{211.97} \\
  \hline
  \makecell{PCh} & \makecell{\#} & \makecell{512} & \makecell{195.5*} & \makecell{191.64} \\
  \hline
  \makecell{REG} & \makecell{\#} & \makecell{256} & \makecell{63.84} & \makecell{1408} \\
  \hline
  \makecell{SRAM} & \makecell{\# words\\\# bits} & \makecell{256\\32} & \makecell{0.93} & \makecell{737.28} \\
  \hline
  \makecell{\textbf{Cores}} & \makecell{\textbf{\#}} & \makecell{\textbf{4096}} & \makecell{\textbf{17E6}} & \makecell{\textbf{152E6}} \\
  \hline
  \hline
  \multicolumn{5}{c}{\textbf{Router}}\\
  \hline
  \hline
  \makecell{MUX IN} & \makecell{\# ports} & \makecell{4} & \makecell{4.8} & \makecell{2.76} \\ 
  \hline
  \makecell{MUX OUT} & \makecell{\# ports} & \makecell{4} & \makecell{4.8} & \makecell{2.76} \\ 
  \hline
  \makecell{REG} & \makecell{\#} & \makecell{2} & \makecell{0.46} & \makecell{11} \\
  \hline
  \makecell{Buffer IN} & \makecell{\# words\\\# bits} & \makecell{4\\32} & \makecell{663.62} & \makecell{704} \\
  \hline
  \makecell{Buffer OUT} & \makecell{\# words\\\# bits} & \makecell{4\\32} & \makecell{663.62} & \makecell{704} \\
  \hline
  \makecell{Accumulator} & \makecell{\# bits} & \makecell{32} & \makecell{120.77} & \makecell{44.23} \\
  \hline
  \makecell{\textbf{Routers}} & \makecell{\textbf{\# levels}\\\textbf{\#}} & \makecell{\textbf{6}\\\textbf{1365}} & \makecell{\textbf{1.99E6}} & \makecell{\textbf{2E6}} \\
  \hline
  \hline
  \multicolumn{5}{c}{\textbf{Overall}}\\
  \hline
   &  &  & \thead{\textbf{Power [$W$]}} & \thead{\textbf{Area [$mm^2$]}}\\
  \hline
  \makecell{\textbf{Total}} &  &  & \makecell{\textbf{19}} & \makecell{\textbf{154.21}} \\
  \hline
  \hline
  \end{tabular}
\end{table*}
\end{scriptsize}

\begin{table*}
    \centering
        \caption{Inference throughput and latency for various datasets and accelerators}
    \begin{tabular}{|c|c|c|c|c|c|c|c|c|}
    \hline
         \makecell{\textbf{Dataset}\\\textbf{ID}} & \makecell{\textbf{Latency}\\\textbf{X-TIME [s]}} & \makecell{\textbf{Throughput}\\\textbf{X-TIME [Sa/s]}} & \makecell{\textbf{Latency}\\\textbf{GPU [s]}} & \makecell{\textbf{Throughput}\\\textbf{GPU [Sa/s]}} & \makecell{\textbf{Latency}\\\textbf{Digital ASIC [s]}} & \makecell{\textbf{Throughput}\\\textbf{Digital ASIC [Sa/s]} } & \makecell{\textbf{Latency}\\\textbf{FPGA [s]}} & \makecell{\textbf{Throughput}\\\textbf{FPGA [Sa/s]}}
 \\  \hline
         1 & 83E-9 & 490E6 & 170E-6 & 4.1E6 & 129E-9 & 124E6 & 44E-6 & 498E3 \\  \hline
         2 & 94E-9 & 247E6 & 14E-3  & 3.4E6 & 122E-9 & 31E6  & 247E-6& 83E3\\  \hline
         3 & 127E-9& 62.5E6& 5.4E-3 & 4.7E6 & 155E-9 & 31E6  & 143E-6 & 196E3\\  \hline
         4 & 202E-9& 58.6E6 & 17E-6 & 23.5E6& 225E-9 & 31E6  & 144E-6 & 156E3\\  \hline
         5 & 106E-9& 32.3E6 & 8.6E-3& 4.3E6 & 134E-9 & 31E6 & 200E-6 & 111E3\\  \hline
         6 & 93E-9 & 327E6  & 234E-6 & 128E6 & 136E-9 & 124E6 & 18E-6 & 1.3E6\\  \hline
         7 & 94E-9 & 250E6 & 11E-3 & 4.4E6 & 122E-7 & 31E6 & N/A & N/A \\  \hline
    \end{tabular}
    \label{si:tab:benchmark}
\end{table*}

\begin{table*}
    \centering
        \caption{Mean and standard deviation of the model performance across the datasets for multiple noise profiles}
    \begin{tabular}{|c|c|c|c|c|c|c|c|}
     \hline
         \textbf{Dataset ID} &  \textbf{Metric} & \textbf{ReRAM (mean)} & \textbf{ReRAM (std)} & \textbf{DAC (mean)} & \textbf{DAC (std)} & \textbf{Both (mean)} & \textbf{Both (std)}\\ \hline
        1 & Accuracy & 0.86 & 0 & 0.86 & 0 & 0.86 & 81E-6 \\ \hline
        2 & Accuracy &0.73 & 4.5E-6 & 0.73 & 0 & 0.73 & 56E-6 \\ \hline
        3 & Accuracy &0.96 & 13E-6 & 0.96 & 0 & 0.96 & 9E-6 \\ \hline
        4 & Accuracy &0.99 & 2.2E-16 & 0.99 & 0 & 0.99 & 42E-6 \\ \hline
        5 & Accuracy &0.7 & 3.3E-16 & 0.7 & 0 & 0.7 & 60E-6 \\ \hline
        6 & Accuracy &0.82 & 141E-6 & 0.82 & 0 & 0.82 & 125E-6 \\ \hline
        7 & RMS &503.5 & 0.15 & 503.5 & 0 & 506.1 & 0.65 \\ \hline
    \end{tabular}
    \label{si:tab:noise}
\end{table*}

\clearpage
\bibliographystyle{IEEEtran}
\bibliography{supplementary_information}